\pdfoutput=1

\documentclass[11pt]{article}

\usepackage[]{acl}

\usepackage{times}
\usepackage{latexsym}

\usepackage[T1]{fontenc}

\usepackage[utf8]{inputenc}

\usepackage{microtype}

\usepackage{inconsolata}

\usepackage{graphicx}
\usepackage{amsmath}
\usepackage{hyperref}
\usepackage{tabularx}
\newcolumntype{C}{>{\centering\arraybackslash}X}

\usepackage{booktabs}
\usepackage{multirow, multicol}
\usepackage{comment}
\usepackage{float}
\restylefloat{table}
\usepackage{url}

\usepackage{algorithm}
\usepackage{algpseudocode}
\usepackage{arydshln}
\usepackage{array}
\usepackage{makecell}
\usepackage{multirow}
\usepackage{enumitem}
\usepackage{booktabs}

\usepackage{caption}

\newcommand{\baseline}{{\textsc{NoAction}}}
\newcommand{\aplan}{{\textsc{ActionPlan}}}
\newcommand{\aaware}{{\textsc{ActionAware}}}
\newcommand{\copt}{{\textsc{ComplianceOpt}}}
\newcommand{\guideline}{{\textsc{Guideline}}}
\newcommand{\llm}{{\textsc{LLM-prompting}}}

\usepackage{listings, color}
\lstdefinestyle{llmprompt}{
  language=Python,
  basicstyle=\ttfamily,
  frame=single,
  keywordstyle=\color{blue},
  stringstyle=\color{orange},
  showstringspaces=false,
  keepspaces=true,
  escapeinside={<@}{@>},
  rulecolor=\color{black},
  breaklines=true, 
  breakatwhitespace=true, 
  linewidth=\linewidth, 
  basicstyle=\tiny
}

\usepackage{listings, color}
\lstdefinestyle{human_guideline}{
  basicstyle=\ttfamily,
  frame=single,
  keywordstyle=\color{blue},
  stringstyle=\color{orange},
  showstringspaces=false,
  keepspaces=true,
  escapeinside={<@}{@>},
  rulecolor=\color{black},
  breaklines=true, 
  breakatwhitespace=true, 
  linewidth=\linewidth, 
  basicstyle=\tiny
}
\usepackage{xspace,mfirstuc,tabulary}
\usepackage{amsfonts}

\title{Workflow-Guided Response Generation for Task-Oriented Dialogue}

\author{
        {Do June Min}\\ 
        University of Michigan \\  \texttt{dojmin@umich.edu}
        \And  
        {Paloma Sodhi}\\ 
        ASAPP \\  \texttt{psodhi@asapp.com}
        \And
        {Ramya Ramakrishnan}\\ 
        ASAPP \\  \texttt{rramakrishnan@asapp.com}
}

\begin{document}
\maketitle
\begin{abstract}

%
Task-oriented dialogue (TOD) systems aim to achieve specific goals through interactive dialogue.
Such tasks usually involve following specific workflows, i.e. executing a sequence of actions in a particular order.
While prior work has focused on supervised learning methods to condition on past actions, they do not explicitly optimize for compliance to a desired workflow.
In this paper, we propose a novel framework based on reinforcement learning (RL) to generate dialogue responses that are aligned with a given workflow.
Our framework consists of ComplianceScorer, a metric designed to evaluate how well a generated response executes the specified action,
combined with an RL optimization process that utilizes an interactive sampling technique.
We evaluate our approach on two TOD datasets, Action-Based Conversations Dataset (ABCD) \cite{chen2021abcd} and MultiWOZ 2.2 \cite{zang-etal-2020-multiwoz} on a range of automated and human evaluation metrics.
Our findings indicate that our RL-based framework outperforms baselines and is effective at generating responses that both comply with the intended workflows while being expressed in a natural and fluent manner.

\end{abstract}
\begin{figure}[t]
    \centering
    \includegraphics[width=\linewidth]{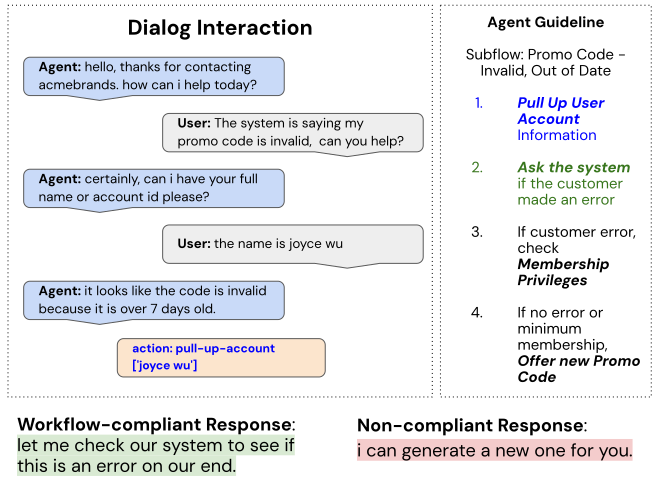}
    \caption{In this interaction, 
    the customer requests assistance with an expired promo code.
    The agent must help the customer while following the steps in
    the agent guideline, which consists of a sequence of actions that must be taken in order to resolve the issue. 
    For example, offering to generate a new promo code without querying the system results in a non-workflow-compliant behavior.
    }
    \label{fig:interaction} 
\end{figure}

\section{Introduction}

Task-oriented dialogue (TOD) focuses on creating conversational systems that assist users in attaining specific objectives. While prior TOD literature has extensively looked at predicting user intents and identifying relevant slots and values \cite{henderson-etal-2014-second, wei-etal-2018-airdialogue,  budzianowski-etal-2018-multiwoz, byrne-etal-2019-taskmaster, rastogi2020_sgd, shalyminov2020fast, balaraman-etal-2021-recent}, real-world interactions often involve nuanced workflows, and optimizing for such workflows remains underexplored \cite{chen2021abcd, Hattami2022WorkflowDF, raimondo2023improving}. Consider a customer support interaction where agents must follow multi-step procedures that adhere to company policies. For example, in Figure~\ref{fig:interaction}, a customer asks for help with an expired promotional code. 
A model that accounts for the user intent might respond reactively, offering to generate a new promo code. However, assisting the customer involves not only modeling their intent, but also staying consistent to a workflow, in this case the company policy. This involves the agent executing the necessary actions in the right order, such as pulling up account information and querying the system to make sure the customer qualifies for the promotion.

Many prior approaches in task-oriented dialogue (TOD), such as SimpleTOD \cite{hosseini2020simple} and PPTOD \cite{su-etal-2022-multi}, have employed supervised learning alongside utterance-level user intents and system dialogue acts (DAs) for system response generation. 
However, these frameworks lack explicit optimization for compliance in response generation, resulting in responses that may fail to execute the specified action. 
This problem arises because the response generators neither receive rewards nor penalties based on the adherence to the specified actions. 
Additionally, there is a notable absence of a metric or model to quantitatively assess the degree of compliance, hindering the evaluation and training of response generators.

In this work, we tackle the problem of workflow-compliant response generation in TOD and propose an RL-based approach that addresses the limitations of existing systems.
Our approach (\textsc{\copt}) employs RL with compliance scoring to construct training data for the Quark \cite{lu2022_quark} framework. 
We evaluate our approach using the Action-Based Conversations Dataset (ABCD) \cite{chen2021abcd}, a TOD dataset enriched with policy-based agent behavior constraints in the form of action sequences, and MultiWOZ 2.2 \cite{zang-etal-2020-multiwoz}. 
Our experiments show that models integrating workflow information surpass baseline models, producing responses that adhere to policies while maintaining a natural and fluent tone. 
Furthermore, we observe that direct compliance optimization through RL can lead to additional enhancements in the workflow compliance levels of the dialogue system. 
We validate our results through automated metrics and human evaluations.
Our key contributions include:
\begin{itemize}[nosep, leftmargin=0.2in]
   \item 
   A reinforcement learning (RL)-based framework for training workflow-compliant response generators, based on an interactive sampling technique to optimize model behavior over multiple dialogue exchanges.
   \footnote{We open-source our code at \url{github.com/TODO}.}
   \item A new compliance metric based on a reward model scorer as well as an LLM score and validate it against human evaluations
    \item Evaluation on both automated and human evaluation metrics showing that our models, enhanced with workflow information and direct compliance optimization through RL, consistently outperform baselines.
\end{itemize}


\section{Related Work}

\paragraph{Task-oriented Dialogue.}
Recently, there has been an increase in TOD tasks and datasets \cite{henderson-etal-2014-second, budzianowski-etal-2018-multiwoz, byrne-etal-2019-taskmaster, wei-etal-2018-airdialogue, rastogi2020_sgd},
indicating a growing emphasis on advancing natural language processing techniques for practical applications. 
These datasets encompass diverse domains and enable researchers to tackle a wide spectrum of real-world challenges. 
However, previous benchmarks have predominantly focused on evaluating only some components of TOD systems, such as intent recognition and slot filling, with limited focus on aspects like workflow compliance \cite{chen2021abcd}.

\paragraph{Workflow Compliance.}

The problem of workflow compliance is closely related to, but distinct from, the task of dialogue policy management. 
The primary objective of dialogue policy management 
is to predict the optimal dialogue action based on the current conversation state \cite{takanobu-etal-2019-guided, he2022galaxy}. 
In this context, dialogue actions represent intentions or decisions that are isolated to a single user query, such as ``book a flight'' or ``find a nearby restaurant.'' 
In contrast, workflow compliance adopts a more holistic approach, considering the sequential workflow from the larger context of the conversation to define success. 
For example, offering a new promo code is only valid after a system check has been executed 
 first (Figure~\ref{fig:interaction}).
It emphasizes the fact that user interactions are not isolated actions but rather part of a continuous process with multiple steps. 
\citet{raimondo2023improving} expands upon \citet{chen2021abcd}'s work to show that models augmented with workflow-specific information such as workflow names or action plans can boost the generalizability of action prediction models, but does not consider the problem of generating workflow-compliant responses, which is a focus of our work.

\paragraph{Reinforcement Learning.}
RL has been successfully used to improve TOD systems \cite{pietquin2011_sample,gasic-etal-2013-pomdp, fatemi-etal-2016-policy,lewis-etal-2017-deal,  singh2022_optimizing}.
One application 
is training and improving dialogue managers that maintain dialogue state transitions \cite{rieser_and_lemon_2011}.
Another focus is to use RL in conjunction with supervised learning to improve the quality of language generation, such as in \cite{lewis-etal-2017-deal}.
This line of research applies similar techniques used in RL for general-domain dialogue generation, such as interleaving supervised learning and RL, offline and online RL, policy gradients, and Q-learning  \cite{li-etal-2016-deep, JLK2022,snell2023_ilql,pmlr-v202-sodhi23a}.
Our work adopts a similar strategy of supervised learning followed by RL but introduces a novel interactive sampling technique. 




\section{Problem Formulation}

\subsection{Workflow-Compliant Response Generation as an MDP}
We formalize the problem of workflow-compliant response generation as a Markov Decision Process (MDP).
Given a dataset of context-response pairs $\{ x^i, y^i \}_{i=1}^{N}$, where context $x$ is the conversation history, and response $y = \{y_1, \dots, y_T \}$ is a target sequence of tokens.

Additionally, each dialogue is associated with a domain $d$ representing the task (e.g., troubleshoot-site, subscription-inquiry). 
Every domain has an associated set of workflow $G_d$, which is a natural language description of the steps the system must follow in order to assist the user, as well as a sequence of actions $W_d$, which represents a flat action list or workflow sequence based on the guidelines. Fully compliant dialogues do not necessarily follow the full sequence $W_d$ and may instead only include a subset of these actions since the guideline includes conditional branching. For example, in Figure~\ref{fig:interaction}, the workflow step 3 is dependent on the result of step 2.

Each data instance, denoted as $(x, y, G_d)$, can be viewed as an episode within a MDP, which we define as follows:

\begin{itemize}[nosep]
\item \textbf{States, $s_t \in \mathcal{S}$} is the context $x$, workflow $G_d$, and the partially generated sequence of tokens up to and including time step $t$, which we denote as $\hat{y}_{<t}:={\hat{y}_1,\dots, \hat{y}_t}$.

\item \textbf{Actions, $a_t\in \mathcal{A}$} are the set of possible next tokens $\hat{y}_{t+1}$ from the vocabulary $V$.

\item \textbf{Transition function, $\mathcal{T}(s_{t+1} | s_t, a_t)$} is deterministic , as each state-action pair $(\hat{y}_{{<t}}, \hat{y}_{{t+1}})$ leads to a unique state $\hat{y}_{<t+1}$ for the next step.

\item \textbf{Rewards, $r_t: \mathcal{S}\times\mathcal{A} \rightarrow [0, 1]$} provide a measure of how well the generated response $\hat{y}$ executes the provided workflow $G_d$.
It is a terminal reward. 
When workflow compliance can be computed only after multiple exchanges, the reward is computed using \emph{block evaluation}.


\item \textbf{Horizon, $T$} 
represents the time span of each episode, concluding either when the current time step $t$ exceeds $T$ or when an end-of-sentence (EOS) token is generated.

\end{itemize}

The goal is to learn a policy $\pi: s_t \rightarrow a_t$ maximizing \textit{return}, \emph{i.e.} 
the cumulative reward over an episode $\mathbb{E}_{\pi} \sum_{t=0}^T \gamma^t r_t$. 
We assume undiscounted cumulative rewards, \emph{i.e.} $\gamma=1$.

\paragraph{Block Evaluation.}
One of our key observations is that compliance is not easily captured in a single dialogue response. For example, in a customer service use case, an agent may need to verify the identity of the user before proceeding to issue resolution. 
To successfully be in compliance with the next workflow action e.g. \texttt{verify-identity}, the agent needs to take several steps. In order to better model and leverage this insight, we consider ``blocks'' of user and agent utterances when evaluating and optimizing for compliance. Blocks refer to the sequence of user and system utterances that occur between two action executions. We now continue with detailed descriptions of each of the approaches.
We define an interaction ``block'' $b$ as a list of user and system utterances between consecutive action executions by the system. 



\begin{figure}[!t]
    \centering
    \includegraphics[width=1.0\linewidth]{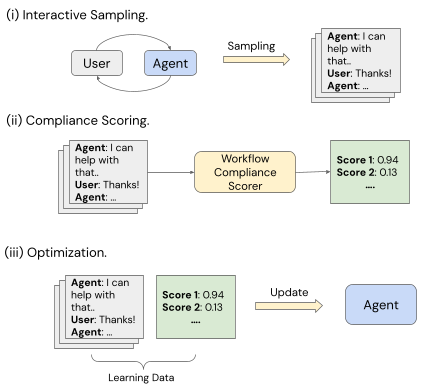}
    \caption{\textbf{Approach Overview.} 
RL optimizes the model towards better workflow compliance. Interaction-score pairs are processed into RL data in the Quark framework.
}
    \label{fig:approach} 
\end{figure}

\section{Approach}
\label{sec:method}
We introduce \textsc{\copt}, which directly optimizes compliance with the specified workflow.
We define compliance as the extent to which the generated system utterances adhere to the prescribed workflow action at turn $t$.

Algorithm~\ref{alg:comptliance_opt_training} shows our overall training procedure using the ComplianceScorer, which is adapted from the Quark framework \cite{lu2022_quark}.
The Quark framework is similar to Decision Transformer in that it treats RL as a sequence modeling problem \cite{chen2021_dt}.
After interactively sampled (Figure~\ref{fig:approach}-(\romannumeral 1)) generations are scored (Figure~\ref{fig:approach}-(\romannumeral 2)), the rewards are quantized to produce reward tokens $r_k$, which are then used to condition the generations during training (Figure~\ref{fig:approach}-(\romannumeral 3)).

\subsection{Interactive Sampling}

Diverging from the Quark method, we implement an interactive sampling step, using two distinct models, a system model, and a user simulator. 
This is because achieving workflow compliance often requires multiple dialogue turns between the system and the user. Consider a customer service agent who needs to gather a user's name, email, and order id to validate their purchase. This is a multi-turn process where the system needs to gather information over multiple dialogue turns of questions and answers.



We start with a warm-start system model trained with standard autoregressive training. 
The user simulator remains fixed during Quark training and is only used for the interactive sampling procedure.
Given a dialogue context $c_0 = [u_0]$, the system model first samples an utterance, which is then concatenated to $c_0$, forming $c_1 = [u_0, s_1]$.
Then, conditioned on $c_1$, the user simulator samples a user turn, forming another context $c_2 = [u_0, s_1, u_2]$.
This process is repeated $M$ times, which is a hyperparameter.
We denote the generated user and system utterances \textit{block} as $b$, which are fed alongside planned workflow actions as inputs to the ComplianceScorer.

Our interactive sampling technique is independent of the Quark framework and can be used as a sampling approach for other policy gradient RL methods, such as proximal policy optimization (PPO) \cite{SchulmanWDRK17}.


\subsection{Compliance Scoring Model}
In order to quantify compliance and use it as a reward model for RL, we developed the ComplianceScorer, which measures the alignment between the generated system utterances and the prescribed workflow action.

\paragraph{Reward modeling.}
We train the ComplianceScorer using the reward modeling loss for comparing two continuations given a prompt \cite{Ouyang0JAWMZASR22}.
\begin{equation}
\small
\resizebox{0.9\hsize}{!}{
$l(\theta) = -\sum_{(p,b_w,b_l) \sim D} \log \left( \sigma \left( r_{\theta}(p,b_w) - r_{\theta}(p,b_l) \right) \right)$
}
\label{eqn:reward_loss}
\end{equation}
$r(p, b)$ represents the scalar output generated by ComplianceScorer (parametrized by $\theta$), when given 
the planned workflow action $p$ and the generated block $b$.
The term $b_w$ denotes the favored choice among the pair of context completions, $b_w$ and $b_l$, in the comparison dataset $D$.

The generated block $b$ can include multiple utterances by both the user and the system.
We found that there are advantages in excluding the dialogue context and solely presenting $b$ to the model due to the following reasons: 
(1) The model can concentrate more effectively on evaluating the text itself rather than being distracted by the typically longer context, 
(2) Constructing negative instances ($b_l$) becomes straightforward by substituting the workflow of the positive instance $b_w$ with an alternative.

\paragraph{Data Collection.}
Our approach for constructing the comparison dataset $D$ is simple.
First, we segment each conversation into multiple blocks $s$, comprising contiguous utterances that are annotated with the same workflow step $p$.
By pairing utterances from different segments with different workflow annotation $p_l$ such that $p\ne p_l$, we efficiently generate $(p, b_w, b_l)$ triplets for the ComplianceScorer dataset. 

\algrenewcommand\algorithmicrequire{\textbf{Input:}}
\algrenewcommand\algorithmicensure{\textbf{Output:}}
\begin{algorithm}[t]
\small
\begin{algorithmic}[1]
\Require{Initial Policy $l_0$, User Simulator $\mu$,
Dialogue Contexts $C$, reward $r(\cdot)$, KL weight $\beta$, number of quantiles $K$, number of interactions $M$, number of train iterations $N$}
\State Make a copy $l_\theta$ of initial policy $l_0$.
\For{iteration = $1, 2, \cdots , N$}
    \For{$c_i \in C$}
    \State Do \textbf{\texttt{interactive\_sample}}($l_\theta$, $\mu$, $M$, $c_i$) to obtain $b_i$.
        \Comment{Interactive Sampling}
    \State Add $c_i, s_i, r(c_i, s_i) $into data pool $\mathcal{D}$ 
    \Comment{Scoring}
    \EndFor
    \State $\bar{\mathcal{D}_i}$ $\leftarrow$ \texttt{\text{quantize}}
    \For{step $= 1, 2, \cdots , M$}
    \State{Draw a batch of data $(c_i, b_i, r_{k_i})$ from quantized data pool $\bar{\mathcal{D}_i}$}
    \State{Compute the objective in Eqn~\ref{eqn:quark_objective} and update policy $\theta$ with gradient descent}  \Comment{Update}
    \EndFor
\EndFor
\end{algorithmic}
\caption{\textsc{\copt} RL Training}\label{alg:comptliance_opt_training}
\end{algorithm}

\subsection{Compliance Optimization}
\label{sec:compliance_optimization}

Finally, the system model is updated according to a combination of the standard LM loss and a KL divergence loss between the updated model and the reference model, shown in Equation~\ref{eqn:quark_objective}.

\begin{equation}
\begin{aligned}
    \max_{\theta} \mathbb{E}_{k \sim \mathcal{U}(1,K)} \mathbb{E}_{(c,b) \sim \mathcal{D}^k} [
    \log l_{\theta}(b | c, r_k) &\\- \beta \sum_{t=1}^{T} \text{KL}\left( l_0(\cdot | b_{<t}, c) \| l_{\theta}(\cdot | b_{<t}, c, r_k) \right) ] 
\end{aligned}
\label{eqn:quark_objective}
\end{equation}

\section{Experimental Setup}

\paragraph{Datasets.} 

We evaluate our approach using two TOD datasets designed for real-world applications, where users aim to accomplish specific tasks through dialogue. Each dataset consists of conversations between two speakers, a system or agent and a user or customer.
\begin{itemize}[nosep, leftmargin=0.2in]
\item \textbf{Action Based Conversations Dataset (ABCD)~\cite{chen2021abcd}}: Contains $\sim$10,000 dialogues between customers and agents and spans 55 intents. The agents have explicit workflows they need to follow according to company guidelines, making it an ideal dataset to evaluate compliance requirements.
\item \textbf{MultiWOZ 2.2~\cite{zang-etal-2020-multiwoz}}: Contains over $\sim$10,000 dialogues spanning multiple domains. 
We designate annotated user intents as workflow actions to be predicted and include agent dialogue acts (DAs) in the context.
\end{itemize}


\paragraph{Evaluation.}
We evaluate the different approaches and baselines on a variety of metrics:
\begin{itemize}[nosep, leftmargin=0.2in]
    \item \textbf{Automated Compliance}
    \begin{itemize}[label=$\ast$,left=0em,labelsep=1em,itemsep=0em,parsep=0em]
        \item Compliance Reward Scorer: We evaluate compliance using our automated ComplianceScorer (Section~\ref{sec:compliance_optimization}).
        \item LLM Scorer: We automatically evaluated compliance using an LLM (prompt in Appendix~\ref{sec:appendix}). 
We used a categorical labeling scheme involving 
two levels:  0 = 'not compliant,' and 1 = 'fully compliant.' 
    \end{itemize}
    \item \textbf{Human Annotations}: For human evaluation, we randomly selected 100 generated outputs from each model (guideline in Appendix~\ref{sec:appendix}). 
An identical categorical labeling scheme (0, 1) was used for both \textit{compliance} and \textit{coherence}, involving three annotators. 
The annotators had access to the complete policy document containing guidelines for all workflow actions.
    \item \textbf{Semantic Similarity and Diversity}
    \begin{itemize}[label=$\ast$,left=0em,labelsep=1em,itemsep=0em,parsep=0em]
        \item 
        Similarity: We measure the similarity between the generated responses with the corresponding human-annotated ground truth. We use widely-used similarity measures such as BLEU, Meteor, BLEURT, and BERTScore
\cite{papineni-etal-2002-bleu,  banerjee-lavie-2005-meteor,sellam-etal-2020-bleurt, zhang_2020_bertscore}. We report the ``Block'' version of each of the metrics computed by taking the max of BertScores between each prediction and target utterance pair over all targets and taking the average over predictions.

        \item 
        Diversity: We measure the diversity of generated system responses using the dist-$3$ metric \cite{li-etal-2016-diversity}.
    \end{itemize}    
    \item \textbf{Workflow Accuracy}: For the {\aplan} and {\copt} models that predict the next workflow step, we report the exact match accuracy of the predicted workflow steps against ground truth.
\end{itemize}

\paragraph{Methods \& Baselines.}



\begin{itemize}[nosep, leftmargin=0.2in]
\item \textbf{\baseline}: A simple model that only sees user and system utterances without access to completed actions or next workflow steps, as done in \citet{pmlr-v202-sodhi23a}.
\item \textbf{\aaware}: 
Action executions are interleaved in the input alongside utterances, allowing the model to understand the history of completed workflow actions in the dialogue context. 
The model may implicitly learn the relationship between workflow policies and agent utterances, enabling the generation of more contextually relevant responses. %
This approach applies supervised learning on dialogue context augmented with prior and future actions, and is similar to \citet{hosseini2020simple}.

\item \textbf{\aplan}: The {\aplan} model goes beyond {\aaware} by explicitly modeling future compliance to workflow policy guidelines. 
It introduces the concept of a "planned" workflow action, representing the next action that must be completed based on the policy. 
This planned action is incorporated into the dialogue context, and the model generates responses that align with the intended workflow. 
This approach treats the planned future workflow action as a latent variable in the generation process, resulting in better workflow compliance in responses.

\item \textbf{\guideline}: \label{sec:guideline}
Instead of relying on completed actions or predicted future workflow actions, the \textsc{Guideline} approach conditions on a fixed "standard" sequence of actions, referred to as the guideline in \cite{chen2021abcd}.


\item \textbf{\llm}: \label{sec:llm}
We use prompting and in-context learning with large language models (LLMs) to explore the option of using natural language policy guidelines as a source of workflow information.
Similar to prior work \cite{Zhang2023SGPTODBT},
our prompt to the LLM consists of instructions that describe the task and task-related text that consists of policy guidelines, example conversations, and the dialogue context $C_t$.
Our LLM-prompting method assumes \textsc{Oracle} next workflow and generates corresponding responses.
We include our prompt in Appendix~\ref{sec:appendix}.

\item \textbf{\textsc{Predicted/Oracle}} variants: 
At test time, our {\aplan} and {\copt} models predict the next workflow action and condition system response generation on this action (referred to as \textsc{Predicted}).
In the \textsc{Oracle} option, we supply the models with ground truth next workflow actions, in order to gauge the upper bound of performance.
Finally, \textbf{{\aplan}\textsc{All Future Oracle}} uses all future remaining workflow steps annotated in the ground truth data.
\end{itemize}

\paragraph{Training.}
Our dialogue system models and user simulators are both initialized with pretrained DistilGPT2 \cite{sanh2019distilbert}, which is a condensed variant of GPT-2 \cite{radford2019language}.  
Tokenization of inputs to system and user models use pre-trained BPE codes \cite{sennrich-etal-2016-neural}.
For the ComplianceScorer model, we start with a pretrained RoBERTa model, with its associated BPE tokenizer.
\cite{liu2019_robeta}.
%
Training hyperparameters are included in Appendix~\ref{sec:appendix}.

\section{Results and Analyses}

\subsection{Overall Results}

\begin{figure*}[t]
\includegraphics[width=1.0\linewidth]{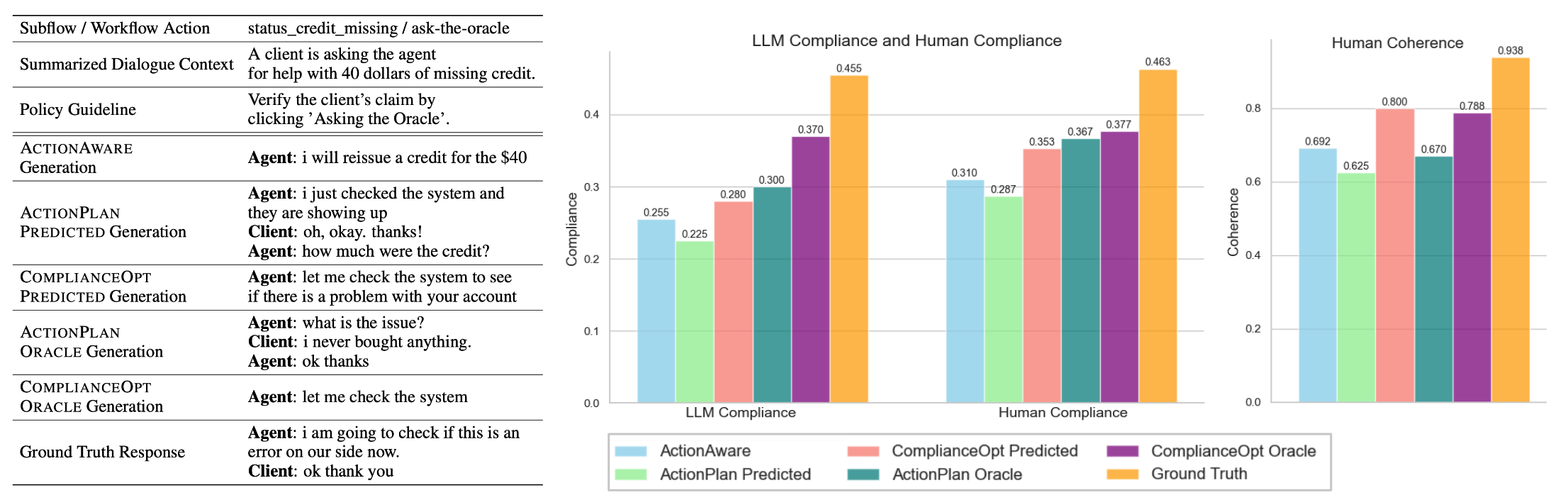}
\captionof{figure}{
\textbf{Left}: Sample Simulated Interaction between Agent Models and User Simulator.
\textbf{Right}: 
Evaluation results with human annotators and LLM.
We report the average score received for each model.
}
\label{fig:bar_chart}
\end{figure*}

\begin{itemize}[nosep, leftmargin=0.2in]
\item \textit{Workflow-awareness consistently improves performance:} Models incorporating workflow information all show higher compliance performance over the {\baseline} baseline (Table~\ref{tab:automated_compliance_results}).

\item \textit{Direct compliance optimization leads to peak system compliance:} Our investigation reveals that \textsc{\copt}, which utilizes reinforcement learning to optimize compliance scores, outperforms models trained with teacher-forcing. 
This approach excels not only in automated compliance scoring (Table~\ref{tab:automated_compliance_results}) but also in human compliance and LLM-based evaluations (Figure~\ref{fig:bar_chart}).

\item \textit{Human evaluation validates automated metrics:} Human evaluators corroborate the results obtained from automated evaluations, confirming that workflow-aware models consistently outperform baselines. Remarkably, RL optimization achieves higher compliance without compromising coherence (Figure~\ref{fig:bar_chart}).

\item \textit{Consistent performance across datasets:} The improved performance of workflow-aware models, particularly \textsc{\aplan} and \textsc{\copt}, extends beyond the primary ABCD dataset. These findings hold true even when validated on more general task-oriented dialogue datasets, such as MultiWoz (Table~\ref{tab:multiwoz_experiment_results}).

\item \textit{Ablation Studies:} Lastly, we conduct ablation studies to investigate the effectiveness of explicitly predicting workflow actions compared to directly following standardized workflow guidelines. Additionally, we explore the impact of predicting and conditioning on future action sequences as opposed to single actions (Tables~\ref{tab:automated_compliance_results},\ref{tab:similarity_diversity_results}).
\end{itemize}

\subsection{Automated Compliance Evaluation}
\begin{table}[!htbp]
\small
\renewcommand{\arraystretch}{1.0}
\centering
\resizebox{0.9\linewidth}{!}{%
\begin{tabular}{cc}
\toprule
Model & Compliance Score \\ \midrule
 \multicolumn{2}{c}{Baselines \& Ablations}\\ \midrule
\baseline & 0.4963 \\ 
 {\guideline} \textsc{Oracle} & 0.5713 \\
{\llm} \textsc{Oracle} & 0.8410 \\
{\aplan} \textsc{All Future Oracle} & 0.6043 \\ \midrule 
 \multicolumn{2}{c}{Proposed Methods}\\ \midrule
 \aaware & 0.6012 \\ \hdashline
{\aplan} \textsc{Predicted} & \textbf{0.6762} \\
{\copt} \textsc{Predicted} & 0.6742 \\\hdashline
{\aplan} \textsc{Oracle} & 0.7925 \\ 
{\copt} \textsc{Oracle} & \textbf{0.8670} \\ \midrule
Ground Truth & 0.8676 \\ \bottomrule
\end{tabular}
}
\caption{Compliance Score Results}
\label{tab:automated_compliance_results}
\end{table}


\begin{table*}[!htbp]
\small
\renewcommand{\arraystretch}{1.0}
\centering
\resizebox{0.85\textwidth}{!}{%
\begin{tabular}{c|cccc|c|c}
\toprule
Model & \begin{tabular}[c]{@{}c@{}}Block\\ BertScore\end{tabular} & \begin{tabular}[c]{@{}c@{}}Block\\ BLEURT\end{tabular} & \begin{tabular}[c]{@{}c@{}}Block\\ METEOR\end{tabular} & \begin{tabular}[c]{@{}c@{}}Block\\ BLEU\end{tabular} & dist-3 & \begin{tabular}[c]{@{}c@{}}Workflow\\ Accuracy\end{tabular} \\ \midrule
\multicolumn{7}{c}{Baselines \& Ablations}\\ \midrule
\baseline & 0.1858 & 0.2286 & 0.0549 & 0.4481 & 0.7738 & N/A \\
{\guideline} \textsc{Oracle} & 0.2266 & 0.2763 & 0.0679 & 0.4928 & 0.7536 & N/A \\
{\llm} \textsc{Oracle} & 0.2634 & 0.3933 & 0.0609 & 0.5493 & 0.7013 & N/A \\
{\aplan} \textsc{All Future Oracle} & 0.1624 & 0.2498 & 0.0535 & 0.4606 & 0.7479 & N/A \\ \midrule
\multicolumn{7}{c}{Proposed Methods}\\ \midrule
\aaware & 0.2321 & 0.2703 & 0.0726 & 0.4745 & 0.7661 & N/A \\\hdashline
{\aplan} \textsc{Predicted} & 0.2557 & 0.2959 & 0.0808 & 0.4951 & \textbf{0.7707} & \textbf{0.7011} \\
{\copt} \textsc{Predicted} & \textbf{0.2930} & \textbf{0.2964} & \textbf{0.0924} & \textbf{0.5075} & 0.6553 & 0.6821 \\\hdashline
{\aplan} \textsc{Oracle} & 0.2698 & 0.3081 & 0.0881 & 0.5021 & \textbf{0.7683} & N/A \\
{\copt} \textsc{Oracle} & \textbf{0.3227} & \textbf{0.3312} & \textbf{0.1156} & \textbf{0.5287} & 0.6591 & N/A \\\midrule
Ground Truth & N/A & N/A & N/A & N/A & 0.7738 & N/A \\ \midrule
\end{tabular}
}
\caption{Semantic Similarity and Diversity Results}
\label{tab:similarity_diversity_results}
\end{table*}

\subsubsection{ComplianceScorer Evaluation}

The results in Table~\ref{tab:automated_compliance_results} demonstrate the effectiveness of our proposed models in achieving high compliance scores. 
Notably, both \textsc{\aplan} and \textsc{\copt} consistently achieve the highest compliance scores among all models, showcasing their ability to generate responses that closely adhere to the ground truth workflows. 
This suggests that incorporating explicit workflow information into the response generation process significantly improves compliance.

We also observe that the compliance score of the \textsc{\aplan} \textsc{Predicted} model is slightly higher than that of the \textsc{\copt} \textsc{Predicted} model, while the opposite is true for the \textsc{Oracle} variants.
The lower compliance score of the {\copt} \textsc{Predicted} model could be attributed to the lower workflow accuracy (68\% vs 70\%, Table~\ref{tab:similarity_diversity_results}), and possible scorer model noise.
Thus, we complement our automatic evaluation with LLM-prompting and human evaluation.

Furthermore, it is worth highlighting the performance comparison between the \textsc{Oracle} and \textsc{Predicted} variants. 
The \textsc{Oracle} variants consistently outperform the \textsc{Predicted} variants for both {\aplan} and {\copt}, indicating that having access to the true workflow information during generation contributes significantly to improved compliance. 
This result confirms the importance of accurate workflow prediction in generating compliant responses.

\subsection{Human Evaluation}

As shown in Figure~\ref{fig:bar_chart}, workflow-aware models continue to outperform alternative models in terms of LLM and human evaluation. {\copt} models are not only more compliant but also more coherent than their {\aplan} counterparts, in both \textsc{Predicted} and \textsc{Oracle} variants.

We compute the correlation between ComplianceScorer judgments and human evaluation (Table~\ref{tab:scorer_evaluation}).
Overall, we find a moderate correlation between the automated compliance scoring model and human judgment, showing that our automated compliance scoring model aligns with human evaluators' assessments.

\subsection{Semantic Similarity and Diversity}


{\aplan} and {\copt} achieve the highest semantic similarity scores when compared with ground truth compliant responses (Table~\ref{tab:similarity_diversity_results}). 
This result indicates that adding future planned actions can lead to more contextually relevant and compliant system responses.
Moreover, {\aplan} and {\copt} \textsc{Oracle} models outperform their \textsc{Predicted} counterparts, which suggests that using the \textit{true} next workflow action results in responses more aligned with the human-annotated compliant behavior.

In addition, the lower {dist-3} scores obtained by the {\copt} models, regardless of whether they are in the \textsc{Predicted} or \textsc{Oracle} configuration, suggest that these models produce responses that are linguistically less diverse.
One possible explanation is that the {\copt} models, as a result of the RL-based optimization, learn to focus on generating a narrower range of utterances that are compliant given the context.
Since the ground truth responses achieve both higher {dist-3} and compliance scores, this effect seems unique to the RL optimization.

\subsection{Ablations}
\paragraph{Effect of including all future actions.}

Since including future workflow actions results in more compliant responses, we explore if adding \textit{all} future workflow actions would result in even more compliant behavior ({\aplan} \textsc{All Future Oracle} vs. {\aplan} \textsc{Oracle}). Table~\ref{tab:automated_compliance_results} shows that including all future actions can hurt performance, likely because too much future information leads to noise and model confusion. 
On the other hand, simply focusing on the next workflow action leads to compliant localized interactions (``blocks''). 

\paragraph{Training a model with standardized workflows.}
We consider the effect of conditioning on standardized workflows, without dynamically including next workflow actions in the context. As shown in Table~\ref{tab:automated_compliance_results}, the {\guideline} \textsc{Oracle} model performs better than the baseline but worse than workflow-aware models because it does not dynamically generate contextually relevant workflow actions and responses. 
This reinforces the importance of dynamic workflow prediction, which captures the inherent uncertainty in dialogues.




\paragraph{Few-shot LLM prompting with workflow guidelines.}
The final model variant we considered was directly using an LLM to predict the next workflow action, instead of fine-tuning a separate model.
The {\llm} \textsc{Oracle} model achieves the second-highest compliance score after the {\copt} \textsc{Oracle}. 
We see that the {\copt} model, explicitly trained to optimize compliance is able to outperform or match the LLM with orders of magnitude more parameters (\texttt{gpt-3.5-turbo}).
The high text similarity scores achieved by the {\llm} \textsc{Oracle}, often outperforming even the best-performing {\aplan} and {\copt} models in terms of metrics like BLEURT and BLEU, validate the value of using guided prompts to improve response compliance.
We note that RL optimization of LLMs requires much larger computational resources and remains as interesting future work.

\subsection{MultiWOZ Experiment Results}

\begin{table}[!htbp]
\Huge
\renewcommand{\arraystretch}{1.0}
\centering
\resizebox{1.\linewidth}{!}{%
\begin{tabular}{c||c|cccc|c}
\toprule
Model                                                                  & \begin{tabular}[c]{@{}c@{}}Compliance\\ Score\end{tabular} & \begin{tabular}[c]{@{}c@{}}  BertScore\end{tabular} & \begin{tabular}[c]{@{}c@{}} BLEURT\end{tabular} & \begin{tabular}[c]{@{}c@{}}METEOR\end{tabular} & \begin{tabular}[c]{@{}c@{}} BLEU\end{tabular} & dist-3        \\ \midrule
\multicolumn{7}{c}{Baseline}\\ \midrule
\baseline & 0.7446 & -2.1336 & 0.2476 & 0.0959 & 0.0108 & 0.4366  \\ \midrule
\multicolumn{7}{c}{Proposed Methods}\\ \midrule
\aaware & 0.8451 & 0.1556 & 0.4001 & 0.1959 & 0.0252 & 0.8086 \\ \hdashline
\begin{tabular}[c]{@{}c@{}}{\aplan} \\\textsc{Predicted} \end{tabular}& 0.8463 & 0.1091 & 0.3928 & \textbf{0.1936} & 0.0249 & 0.8027 \\  
\begin{tabular}[c]{@{}c@{}}{\copt}\\ \textsc{Predicted}\end{tabular} & \textbf{0.8853} & \textbf{0.1801} & \textbf{0.4310} & 0.1917 & \textbf{0.0265} & \textbf{0.8267} \\ \hdashline 
\begin{tabular}[c]{@{}c@{}}{\aplan}\\\textsc{Oracle}\end{tabular} & 0.8573 & 0.1149 & 0.3897 & 0.1900 & 0.0242 & 0.7962 \\ 
\begin{tabular}[c]{@{}c@{}}{\copt} \\\textsc{Oracle}\end{tabular}& \textbf{0.9153} & \textbf{0.1837} &\textbf{ 0.4271} & \textbf{0.1951} & \textbf{0.0278} & \textbf{0.8137} \\ \midrule 
Ground Truth & 0.8946 & N/A & N/A & N/A & N/A & 0.8237  \\ \bottomrule 
\end{tabular}
}
\caption{Automated Evaluation Results on MultiWOZ 2.2.
\textsc{Predicted} variants of {\aplan} and {\copt} achieved 69\% and 75\% workflow accuracy respectively. 
}
\label{tab:multiwoz_experiment_results}
\end{table}

In our MultiWOZ experiments, we find consistent support for our approach. 
Workflow-aware models, particularly {\aplan} and {\copt}, outperform {\baseline} and {\aaware} in both \textsc{Predicted} and \textsc{Oracle} settings, showcasing their capacity to generate compliant and contextually relevant responses. 

However, there are several differences when compared to the ABCD experiments.
MultiWOZ introduces increased response diversity, especially noticeable in the {\copt} models, a departure from the ABCD dataset's behavior. 
Moreover, workflow-aware models benefit significantly from action annotation, as seen in the {\baseline} versus {\aaware} comparison.
We conjecture that these disparities may be attributed to differences in action annotation and the nature of actions, which are typically resolved in a single interaction in MultiWOZ, in contrast to the more intricate workflows in the ABCD dataset.

\section{Conclusion}
In this paper, we propose the problem of workflow-guided response generation and introduce a novel RL-based framework to train workflow-compliant models for task-oriented dialogue. 
By integrating workflow information during training and directly optimizing for compliance, our approach improves upon baseline models and generates responses that are both workflow-compliant and linguistically natural. 
We evaluate on both ABCD and MultiWoz datasets and show empirical improvements on automated and human evaluation metrics.


\section{Limitation}
Although we conducted evaluations using automated metrics and human judgments, our evaluation focused on the block-level compliance of generated responses. 
We acknowledge that fully simulating real-world customer service interactions would provide a more comprehensive evaluation of our proposed approaches.
Additionally, 
our user model was intentionally kept simple to facilitate the development and testing of our compliance-driven response generation approaches. 
Using more sophisticated user simulation models that incorporate diverse user behaviors could potentially provide a more realistic assessment of the generalizability of our techniques.
\bibliography{custom, anthology}
\bibliographystyle{acl_natbib}


\appendix
\section{Appendix}\label{sec:appendix}

\subsection{Comparison of the ComplianceScorer with human judgment}
\begin{table}[!htbp]
\large
\centering
\resizebox{\linewidth}{!}{%
\Huge
\begin{tabular}{c||cccc}
\toprule
Model                                                                  & \begin{tabular}[c]{@{}c@{}}\aaware\end{tabular} & \begin{tabular}[c]{@{}c@{}}\textsc{\aplan}\\\textsc{Predicted} \end{tabular} & \begin{tabular}[c]{@{}c@{}}\textsc{\copt}\\\textsc{Predicted} \end{tabular} & \begin{tabular}[c]{@{}c@{}}Ground Truth\end{tabular}  \\ \midrule
Annotator                                                                  &     0.31                                                       &           0.2867                                                 &        0.3533                                                &             0.4633                                            \\ 
Model                                                                  &     0.4349                                                       &           0.5416                                                 &        0.6394                                                &             {0.6290}                                             \\ 
Pearson                                                              &     0.4075                                                       &           0.2114                                                 &        0.2819                                                &             {0.2414}                                             \\ 
Spearman                                                                 &     0.4146                                                       &           0.1645                                                 &        0.2702                                                &             {0.2437}                                             \\ \bottomrule
\end{tabular}
}
\caption{Comparison of the ComplianceScorer with human judgment.
All correlations are significant with $p < 0.05$.
}
\label{tab:scorer_evaluation}
\end{table}


\subsubsection{LLM Compliance Evaluation}

The results, shown in Figure~\ref{fig:bar_chart}, indicate that workflow-aware models outperform baselines on LLM compliance evaluation. {\copt} also achieves higher compliance scores compared with {\aplan}, highlighting the benefit of directly optimizing for compliance.

\subsection{User Simulator}
We instantiate the user simulator with the {\baseline} model.

\subsection{Experiment Details}
We list the parameters and hyperparameters we used for our experiments in Table~\ref{tab:experiment_parameters}.


\begin{table}[!htbp]
\centering
\small
\resizebox{\linewidth}{!}{%
\begin{tabular}{l||c}
\toprule
\multicolumn{2}{c}{Teacher-Forcing Setting} \\
\midrule
Agent model & \texttt{distilgpt2} \\
LLM prompting model & \texttt{gpt-3.5-turbo} \\
Scoring model detail & \texttt{roberta-base} \\
Training epochs & 10 (ABCD) / 1 (MultiWOZ)\\
Learning Rate & 2e-5 \\
Special tokens & START\_USER, 
 START\_WORKFLOW, \\
& END\_WORKFLOW,  START\_AGENT,\\
& END\_AGENT, START\_ACTION, \\
& END\_ACTION, START\_DIALOG, \\
& END\_DIALOGUE \\
\midrule
\multicolumn{2}{c}{Compliance Optimization RL Setting} \\
\midrule
Agent model & \texttt{distilgpt2} (Warm-start {\aplan}) \\
Client model & \texttt{distilgpt2} \\
Learning Rate & 2e-5 \\
Sampling Temperature & 0.5 \\
Number of Interactions & 3 \\
Number of Quantiles $K$ & 5 \\
KL weight $\beta$ & 0.05 \\
Training Steps & 80k (ABCD) / 160k (MultiWOZ) \\
\bottomrule
\end{tabular}
}
\caption{Experiment Models \& Parameters}
\label{tab:experiment_parameters}
\end{table}

\subsection{LLM Prompts}
\begin{lstlisting}[style=llmprompt]
generation_prompt = f"You are a cusotmer agent helping a customer with a issue. Read the dialogue context, provided policy guideline, and generate an agent utterance to help the customer in a way that is compliant to the guideline. The generated agent turn should be at most 2 utterances, and should be similar in length to the agent utterances shown in the examples that demonsrtate compliant agent behavior.\n\Custome Situation: {s}\n\Policy Action Name: {w}\n\Policy Name Guideline: {g}\n\n\{example_str}dialogue Context: {i}\n\n\Agent: "      
\end{lstlisting}


\begin{lstlisting}[style=llmprompt]
evaluation_prompt = f"Read the provide guideline and assess the extent to which the agent's behavior in the input interaction aligns with the specified workflow action, considering the name and a concise description of the workflow provided. 1 = Compliant\n0 = Non-compliant\n\nSubflow: {s}\nWorkflow: {w}\nDescription: {g}\n\n\Dialogue History:\n{i}\n\nInput Interaction:\n{r}\n\nAnswer:"
\end{lstlisting}


\subsection{Human Evaluation Guidelines}
{
\tiny
\paragraph{Compliance:} Assess if the agent's behavior aligns with the specified workflow action, taking into account the action's name and policy guideline. 
If the agent has already completed certain steps or the entire policy guideline behavior in the dialogue history, they should not be penalized for not repeating those corresponding steps.

\paragraph{Coherence:} Rate the coherence of the agent's interaction on a binary scale 
(0=not coherent, 1=coherent). 
In this evaluation, please do not consider repetitive agents as coherent. Additionally, do not include incoherent or disfluent client behavior in the evaluation (only evaluate agent behavior). 
}

\end{document}